\documentclass[runningheads]{llncs}
\usepackage{graphicx}
\usepackage{comment}
\usepackage{amsmath,amssymb} 
\usepackage{color}
\usepackage{subcaption}
\newcommand\blfootnote[1]{%
  \begingroup
  \renewcommand\thefootnote{}\footnote{#1}%
  \addtocounter{footnote}{-1}%
  \endgroup
}

\begin{document}
\pagestyle{headings}
\mainmatter
\def\ECCVSubNumber{3066}  

\title{ Transformation Consistency Regularization-- \\ A Semi-Supervised Paradigm for Image-to-Image Translation} 

\titlerunning{Transformation Consistency Regularization}


\author{Aamir Mustafa  \and
Rafa\l{} K. Mantiuk  }

\authorrunning{A. Mustafa et al.}
%

\institute{Department of Computer Science and Technology, University of Cambridge, UK \\
\email{\{am2806,rafal.mantiuk\}@cl.cam.ac.uk}\\}
\maketitle

\begin{abstract}

Scarcity of labeled data has motivated the development of semi-supervised learning methods, which learn from large portions of unlabeled data alongside a few labeled samples. Consistency Regularization between model's predictions under different input perturbations, particularly has shown to provide state-of-the art results in a semi-supervised framework. However, most of these method have been limited to classification and segmentation applications. We propose \textit{Transformation Consistency Regularization}, which delves into a more challenging setting of image-to-image translation, which remains unexplored by semi-supervised algorithms. The method introduces a diverse set of geometric transformations and enforces the model's predictions for unlabeled data to be invariant to those transformations. We evaluate the efficacy of our algorithm on three different applications: image colorization, denoising and super-resolution. Our method is significantly data efficient, requiring only around 10 -- 20\% of labeled samples to achieve similar image reconstructions to its fully-supervised counterpart. Furthermore, we show the effectiveness of our method in video processing applications, where knowledge from a few frames can be leveraged to enhance the quality of the rest of the movie.

\end{abstract}

\section{Introduction}
\label{introduction}

\blfootnote{Codes are made public at \url{https://github.com/aamir-mustafa/Transformation-CR}}
In recent past, deep neural networks have achieved immense success in a wide range of computer vision applications, including image and video recognition \cite{he2016deep,krizhevsky2012imagenet,karpathy2014large}, object detection \cite{girshick2015fast,ren2015faster}, semantic segmentation \cite{long2015fully,chen2018deeplab} and image-to-image (I2I) translation \cite{isola2017image,zhang2017beyond,dong2015image}. However, a fundamental weakness of the existing networks is that they owe much of this success to large collections of labeled datasets. In real-world scenarios creating these extensive datasets is expensive requiring time-consuming human labeling, e.g. expert annotators, as in case of medical predictions and artistic reconstructions. As we enter the age of deep learning, wide-spread deployment of such models is still constrained for many practical applications due to lack of time, expertise and financial resources required to create voluminous labeled datasets.




Conceptually situated between supervised and unsupervised learning, Semi-Supervised Learning (SSL) \cite{chapelle2009semi} aims at addressing this weakness by leveraging large amounts of unlabeled data available alongside smaller sets of labeled data to provide improved predictive performance. Lately extensive research has been done in SSL and has shown to work well in the domain of image \cite{berthelot2019remixmatch,berthelot2019mixmatch,miyato2018virtual,sajjadi2016regularization,sohn2020fixmatch} and text classification \cite{xie2019unsupervised}. However, it would be highly desirable to create I2I translation networks that can take advantage of the abundance of unlabeled data while requiring only a very small portion of the data to be labeled. For example, to colorize a black and white movie, we may want an artist to colorize only 1--5\% of the frames and rest is done by the network. For capturing video in low light, we may want to capture a few reference frames on a tripod with long exposure times (therefore low noise) and use those to remove noise from the rest of the video. We may also want to design a camera, which captures only every $n$-th frame at a higher resolution (as the sensor bandwidth is constrained) and use those frames to enhance the resolution of the rest of the video. Unsupervised I2I translation methods have shown to generate compelling results, however, although unpaired, they still require large datasets from both the input and output domains to train the network \cite{liu2017unsupervised,zhu2017unpaired,huang2018multimodal,dong2017unsupervised,royer2020xgan,mejjati2018unsupervised,yi2017dualgan}. For example to train an unsupervised super-resolution model we still require huge amounts of high resolution images as in  \cite{yuan2018unsupervised,bulat2018learn}. On the contrary an SSL method would require only low-resolution images and a few low-high resolution image pairs for training. 



In this work, we draw insights from Consistency Regularization (CR) -- that has shown state-of-the art performance in classification tasks -- to leverage unlabeled data in a semi-supervised fashion in a more challenging setting i.e. image-to-image translation. CR enforces a model's prediction to remain unchanged for an unsupervised sample when the input sample is perturbed \cite{bachman2014learning,sajjadi2016regularization,laine2016temporal}. However, applying CR in I2I domain is not straightforward, because images with varied transformations should have different predictions, unlike in the case of image classification. We derive our motivation for our approach from  \textit{a) smoothness assumption}, which in the case of image classification states that if two sample points are close enough to each other in the input space, then their predicted labels must be same and \textit{b) manifold assumption}, which states that natural images lie on a low-dimensional manifold \cite{zhu2009introduction}. 

This paper introduces a regularization term over the unsupervised data called \emph{Transformation Consistency Regularization} (TCR), which makes sure the model's prediction for a geometric transform of an image sample is consistent with the geometric transform of the model's reconstruction of the said image. In other words, we propose a modification to the \textit{smoothness assumption} \cite{zhu2009introduction} for image reconstruction models postulating that if two input images $x, \acute{x} \in \mathcal{X}$ that are geometric transformations of each other and close by in the input space, then their corresponding predictions  $y, \acute{y} $ must be equally spaced in the output manifold. Our training objective is inspired by the \textit{transformation invariance loss} \cite{eilertsen2019single} introduced to design stable CNNs aiming at removing temporal inconsistencies and artifacts in video sequences. Our method is, however, fundamentally different, since the proposed loss formulation and the notion of using unlabeled data in designing near perfect mapping functions (CNNs) have not been previously explored for image-to-image translation settings. The proposed \emph{TCR} over unlabeled data works jointly with the model's supervised training to effectively reconstruct better and visually appealing images than its supervised counterpart. To the best of our knowledge this is the first work to study the applicability of semi-supervised learning in an image-to-image setting.


The main contributions of this paper are as follows:
\begin{enumerate}
    \item Through extensive empirical evaluations, we show the efficacy of our semi-supervised training scheme in regularizing model's predictions to be invariant to input transformations for three image-to-image translation problems, viz. image colorization, denoising and single image super-resolution. 
    \item We hypothesize that addition of unsupervised data during training makes a model generic enough to better remap images from one image manifold to other. We provide validation for our manifold assumption (see Sec.~\ref{sec:validation-manifold}).

    \item We provide analysis of how much unsupervised data is ideal for training a semi-supervised model in contrast to its supervised counterpart per batch. 
    
    \item Using less than 1\% of labeled data, we perform colorization, denoising and super-resolution of movie clips. Our semi-supervised scheme achieves an absolute average gain of 6\,dB in PSNR than its supervised counterpart, which uses the same percentage of labeled data. 
\end{enumerate}

\begin{figure*}[tp]
    \centering
   {\includegraphics[trim={0cm 3.2cm 0cm 0cm}, clip, width=0.95\linewidth]{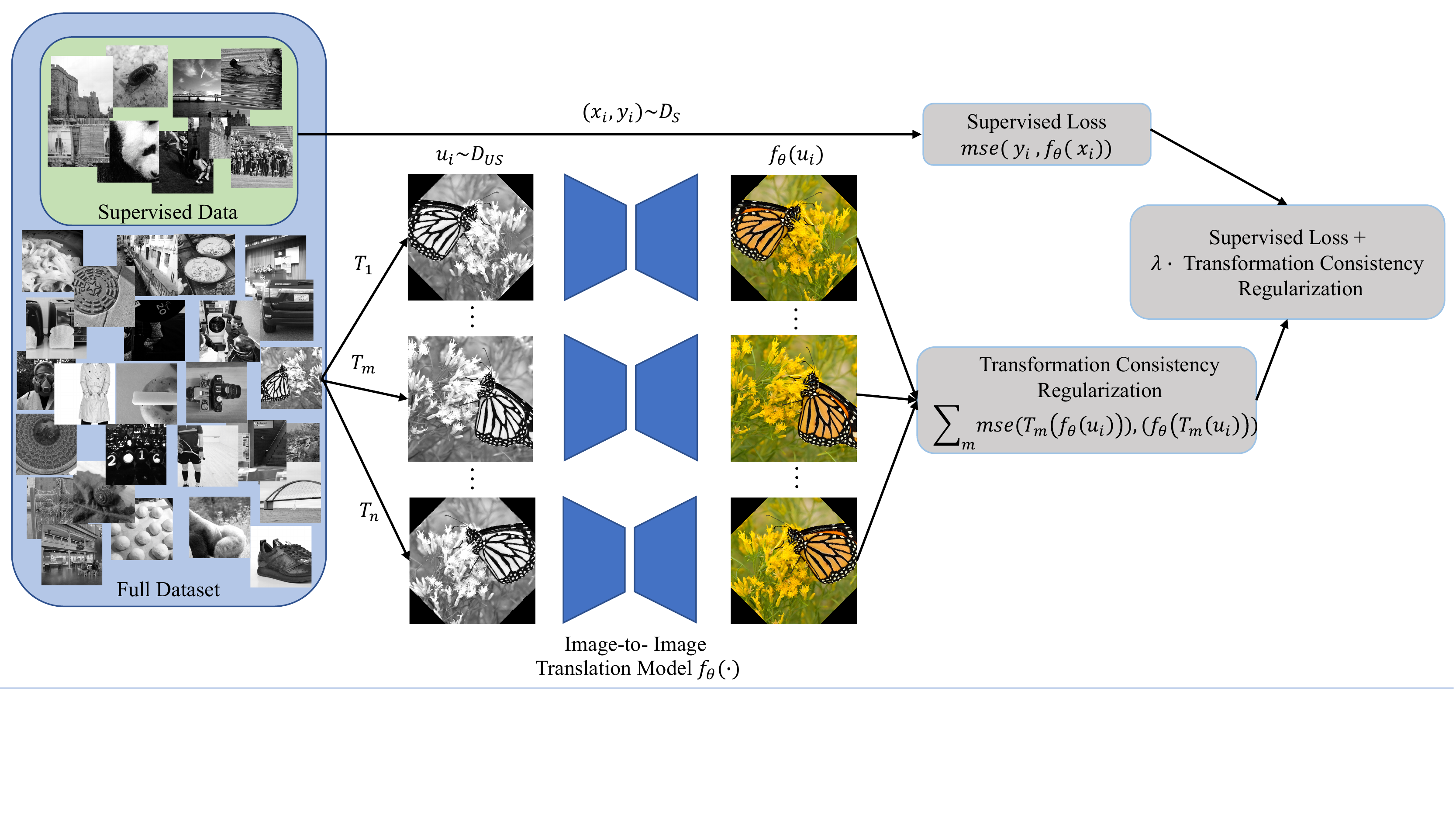} }
    \caption{\small{An illustration of our training scheme using the Transformation Consistency Regularization ($\mathcal{L}_{us}$) over the unlabeled data for image colorization. The same method is used for Image Denoising and Single Image Super Resolution.}}
    \label{fig:block_diag}%
\end{figure*}

\section{Related Work}
\label{sec:related-work}

To set the stage for \textit{Transformation Consistency Regularization} we introduce related existing methods for SSL, particularly focusing on the class of methods that propose addition of an additional loss term while training the neural network, leaving the scheme for training on supervised data unchanged. \textit{Consistency Regularization} (CR) is one such method which enforces the model's prediction to be consistent if realistic perturbations are added to the input data points \cite{athiwaratkun2018there,xie2019unsupervised,miyato2018virtual,sajjadi2016regularization,clark2018semi}. This involves minimizing $d(f_\theta(u), f_\theta(\acute{u}))$, where $\acute{u}$ is a perturbed counterpart of an unlabeled sample $u$ and $d(\cdot,\cdot)$ is a distance metric that measures the offset between the model's predictions. Generally the distance measure used in various works is mean squared error \cite{laine2016temporal,berthelot2019mixmatch}, Kullback-Leibler (KL) divergence \cite{xie2019unsupervised} or the cross-entropy loss. This simple method leverages the unlabeled data to find a low-dimensional manifold on which the dataset lies and has shown to provide state-of-the art performance in image classification tasks. Different CR techniques choose different forms of perturbations added to unlabeled data, most common form include domain-specific data augmentation \cite{xie2019unsupervised,sajjadi2016regularization,berthelot2019mixmatch,laine2016temporal}. More recently, Unsupervised Data Augmentation (UDA) \cite{xie2019unsupervised}, ReMixMatch \cite{berthelot2019remixmatch} and FixMatch \cite{sohn2020fixmatch} enforce consistency among strongly augmented image samples making use of artificial labels generated for weakly augmented samples. Virtual Adversarial Training \cite{miyato2018virtual} adds small amounts of input noise a.k.a adversarial perturbations, which are carefully crafted to significantly alter the models predictions. \cite{laine2016temporal,tarvainen2017mean,park2018adversarial} used dropout noise as a noise injection module to enforce consistency between the model's output predictions. Grandvalet \textit{et al.} introduced \textit{Entropy Minimization} \cite{grandvalet2005semi} making use of unlabeled data to ensure that classes are well separated by adding an additional loss term making sure the model outputs confident (low-entropy) predictions for unsupervised samples. Interpolation Consistency Training \cite{verma2019interpolation} builds on this idea and enforces the predictions at an interpolation of unlabeled images to be consistent with the interpolation of the model's predictions for those images. This achieves the decision boundary to lie on low density regions of the class distribution. MixMatch \cite{berthelot2019mixmatch} combined the ideas of entropy minimization and consistency regularization for better generalization. \textit{Self training} is another such technique that uses the labeled data to initially train a model, which is then used to generate pseudo-labels for unlabeled data. If the model's prediction for an unlabeled sample ($u$) is above a certain threshold, then the sample and its pseudo-label are added to the training set as supervised samples \cite{wang2018towards,chapelle2009semi,nguyen2019semi,zhai2019s4l}. However, the method relies heavily on the threshold value; a larger value results in a small set of pseudo-labeled samples, preventing model's performance to reach its full potential, whereas a smaller threshold may harm the performance with significant amount of erroneous labels. Most of these noise injection methods, however, are designed for image classification problems, with very little work being done in more challenging settings like image-to-image translation.

We in our work broaden the scope of CR to image-to-image translation, which remains untouched by current SSL algorithms. Our aforementioned regularization term and geometric transformations bear closest resemblance to Eilertsen \textit{et al's} \cite{eilertsen2019single} work on colorization of video sequences in designing temporally stable CNNs. They enforce temporal stability by regularizing the various types of motion artifacts that occur between frames in a fully supervised fashion. Our method is, however, fundamentally different, since the proposed loss formulation and the notion of using unlabeled data in designing near perfect mapping functions (CNNs) have not been previously explored for I2I translation settings.


\section{Our Approach}
\label{sec:method}

\subsection{Fully Supervised Perspective}
In a traditional supervised image-to-image learning protocol, we are provided with a finite collection of $B$ image pairs $ D_s = \{ (x_i , y_i) : i \in (1, \ldots, B)\}$  per batch, where each data point $ x_i \in \mathcal{X}$ is sampled from an input distribution $\mathcal{X}$ and $y_i \in \mathcal{Y}$ belongs to a separate target space $\mathcal{Y}$. The goal is to train a regression model, say a Convolutional Neural Network (CNN) $f_{\theta}(\cdot)$ parameterized by $\theta$, which promotes an accurate mapping between the input images $x$ and the ground truth images $y$ by minimizing the loss: 
\begin{equation}
\sum_{i} \mathcal{L} \, \Big( \, f_\theta (x_i) \,, \, y_i \, \Big)
\end{equation}
The loss used to train a CNN could be mean squared error ($L_2$), $L_1$ loss or perceptual loss based on the main objective of the network. 

\subsection{Transformation Consistency Regularization}

In semi-supervised learning, we are provided with an additional set of data points, sampled from the same input distribution $\mathcal{X}$. Let  $D_{us} =\{ (u_i) : i \in (1, \ldots, rB)\}$ be a batch of $rB$ unlabeled data, where $r$ is the ratio of unlabeled to labeled data per batch used in training and is a hyper-parameter.

\begin{figure*}[t]
    \centering
   {\includegraphics[trim={1.3cm 2.4cm 1.1cm 0cm}, clip, width=0.95\linewidth]{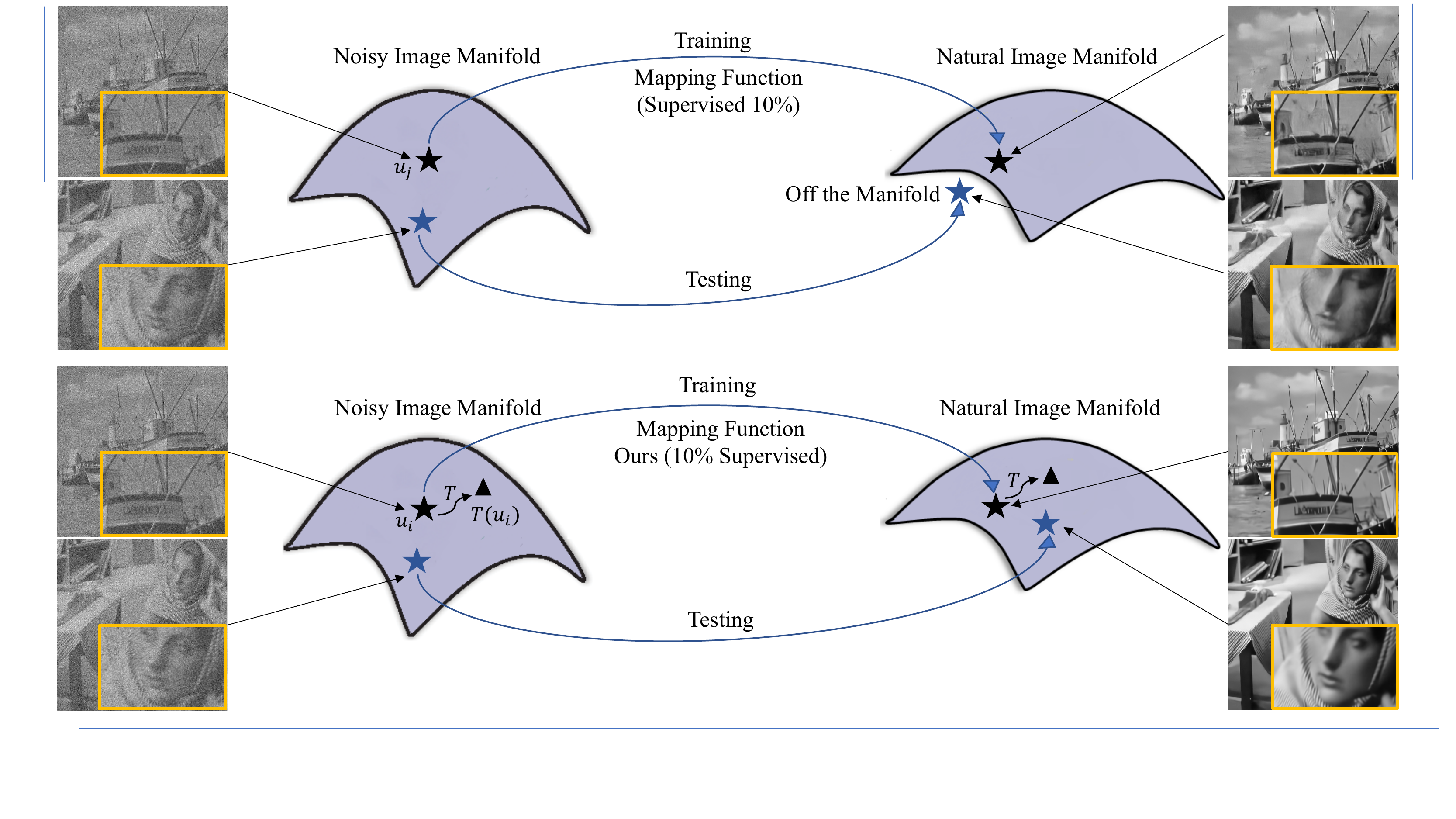} }
    \caption{\small{ The figure illustrates the mapping of image samples from noisy to original image manifold while training and testing. Model trained using only labeled data provides imperfect mapping resulting in reconstructed images lying off-the natural image manifold. Making use of large chunks of unlabeled data for training the same underlying model can provide a better mapping function. Best viewed when zoomed.}}
    \label{fig:image-mapping}
    \vspace{-0.5cm}    
\end{figure*}

Our goal is to leverage the unsupervised data to learn more about the inherent structure of this distribution and thereby regularize the image-mapping network $f_\theta(\cdot)$. We propose a modification to the \textit{smoothness assumption}, postulating that if two input images $x, \acute{x} \in \mathcal{X}$ are close enough in the input space, then their corresponding predictions  $y, \acute{y} $ must be equally spaced. Our approach is also motivated by \textit{manifold assumption} \cite{zhu2009introduction}, which states that natural images lie on a low-dimensional manifold. These assumptions form the basis of semi-supervised learning algorithms in image classification domain  \cite{survey_ssl}. If natural images lie on a low-dimensional manifold, this explains why low-dimensional intermediate feature representations of deep networks accurately capture the structure of real image datasets. Image-to-image translation models, such as an image denoising network, approximate the input image manifold and help in remapping them to the output natural image manifold. Fig.~\ref{fig:image-mapping} shows a low dimensional manifold of noisy and recovered images. Noisy image samples of real world datasets sampled from a distribution $\mathcal{X}$, can be considered to lie on a separate manifold. A trained image denoising model learns to map these data samples to an output natural image manifold. However, insufficient amount of data results in some output images that lie off the manifold. In this work we propose that using additional unlabeled images can go a long way in learning a near perfect mapping function resulting in remapping \textit{off-the-manifold} output images. We provide a detailed validation for our proposition in Sec.~\ref{sec:validation-manifold} 

The foundation of our semi-supervised learning algorithm are the following propositions:

\vspace{1em}
\noindent \textbf{Proposition 1:} \textit{Since both labeled and unlabeled data are sampled from the same underlying data distribution $p(x)$ over the input space, unlabeled data can be used to extract information about $p(x)$ and thereby about the posterior distribution $p(y|x)$.}

\vspace{1em}
\noindent \textbf{Proposition 2:} \textit{The unsupervised data $D_{us}$ provides additional insights about the shape of the data manifold, which results in better reconstruction of images in an image-to-image setting.}

\vspace{1em}
\noindent \textbf{Proposition 3:} \textit{In an image-to-image translation setting, for a diverse set of geometric transformations $T(\cdot)$, if a model predicts a reconstructed image $\hat{u}_i$ for an $i^{th}$ unlabeled input sample $u_i \in D_{us}$, then the model's prediction for its transformation $T(u_i)$ must be $T(\hat{u}_i)$.}

\vspace{1em}

In other words, we propose that a model's output predictions in an image-to-image setting should be transformed by the same amount as the input image is transformed before being fed to the CNN. The loss function for our semi-supervised learning protocol includes two terms \textit{a)} a \emph{supervised loss} term $\mathcal{L}_s$ applied on the labeled data $D_{s}$ and \textit{b)} a \emph{transformation consistency regularization} (TCR) term $\mathcal{L}_{us}$ applied on the combination of $D_{s}$ and $D_{us}$.\footnote{We include all labeled data, without using their labels, alongside the unlabeled data in our transformation consistency regularization term.} For illustration see Fig.~\ref{fig:block_diag}. Specifically, $\mathcal{L}_s$ is the mean-squared loss on the labeled examples:

\begin{equation}
\label{eq:supervised}
\mathcal{L}_{s}(x,y)= \frac{1}{B} \, \sum_{i=1}^{B} \, {\parallel} \, f_\theta (x_i) - y_i \,{\parallel}_2^2
\end{equation}

During the training process, each mini-batch includes both labeled and unlabeled images. The labeled data is trained using a typical I2I translation approach. Additionally, we introduce TCR on both labeled as well as the unlabeled data. We make use of a series of geometric transformations $T (\cdot)$ to enforce consistency between the reconstructed images for a given unlabeled data sample $u_i$. Specifically our loss function is given by:

\begin{equation}
\label{eq:unsupervised}
\mathcal{L}_{us}(u)= \frac{1}{rB} \, \sum_{i=1}^{rB} \, \Big( \frac{1}{M} \, \sum_{m=1}^{M} \,{\parallel}\,  T_m(f_\theta (u_i)) - f_\theta (T_m(u_i)) \,{\parallel}_2^2 \, \Big)
\end{equation}
Here $f_\theta (\cdot)$ is a parametric family of I2I mappings and is a non-linear function, therefore $ T_m(f_\theta (u_i))$ and $f_\theta (T_m(u_i)) $ will indeed have different values. Here $r$ is the ratio of amount of unlabeled data per mini-batch to that of labeled data. The loss function in Eq.~\ref{eq:unsupervised} leverages unsupervised data and thereby helps regularize the model's predictions over varied forms of geometric transformations. The loss in Eq.~\ref{eq:unsupervised} can appear similar to data augmentation with a random transformation but it should be noted that it is fundamentally different. The loss in data augmentation is measured between the outputs and their corresponding ground truths and is typically expressed as ${\parallel}\,T_m(f_\theta (x_i)) - (T_m(y_i))\,{\parallel}$. TCR loss on the other hand enforces consistency between the predictions of transformed inputs without any knowledge of the ground truth.

The overall loss is composed of the original objective of the CNN and our transformation consistency regularization term as:
\begin{equation}
\label{eq:overall}
\mathcal{L}= \mathcal{L}_{s}(x,y) + \lambda \, \Big( \, \mathcal{L}_{us}(u) + \mathcal{L}_{us}(x) \, \Big)
\end{equation}
The scalar $\lambda$ is to control the weight given to the regularization term. Our proposed semi-supervised approach is, however, not restricted to $L_2$ loss and works equally well for $L_1$ and perceptual loss \cite{gatys2015texture,johnson2016perceptual} (see appendix).

\subsection{Transformations}
\label{sec:transformations}

In a traditional consistency regularization (CR) framework for image classification, a noise operation is employed that adds perturbations to the input image $u$ to generate its counterpart $\acute{u}$. An additional regularization term then aims at minimizing the $L_2$ loss, cross entropy loss or KL divergence between the output predictions $f_\theta(u)$ and $f_\theta(\acute{u})$ \cite{laine2016temporal,miyato2018virtual,tarvainen2017mean,xie2019unsupervised}. These noise injection methods include random augmentations, adversarial perturbations, Gaussian noise, dropout noise etc \cite{cubuk2019autoaugment,cubuk2019randaugment}. CR has shown to provide state-of-the art performance in classification domain by enforcing that $u$ and $\acute{u}$ lie on the same image manifold \cite{oliver2018realistic}. 

However, choosing the noise operations in an image-to-image translation model is a challenging task as the output predictions for an unsupervised image sample and, say, its augmented version are no longer same. In this paper we generate unsupervised image pairs using a series of geometric transformations $T(\cdot)$ that capture the occurrences in real world scenarios. These include rotation, translation, scaling and horizontal flipping. We choose a specified range for the degree of transformations so as to prevent images going off their natural manifold. The degree of rotation is uniformly drawn from a range of values between $-45^\circ$ and $45^\circ$. For translation and zooming the range is set between $-30$ px and $30$ px and between $0.9\times$ and $1.1\times$ respectively.

For a particular geometric transformation $T_m$, unlike image classification, the model's predictions for an input image pair $u_i$ and $T_m({u}_i)$ require modifications to be equivalent. The additional loss term for TCR is computed between the model's prediction for $T_m({u}_i)$, denoted by $f_\theta(T_m({u}_i))$  and the transformation of model's output for $u_i$, i.e. $T_m(f_\theta(u_i))$. These transformations over unsupervised data regularize the model and force the reconstructed images to lie on the output image manifold (see Fig.~\ref{fig:image-mapping}).

\section{Experiments}
\label{sec:experiments}

In this section, we evaluate the efficacy of TCR on a variety of image-to-image translation tasks where CNNs are employed. Specifically, we perform experiments with varying amounts of labeled data for image colorization, image denoising and single image super-resolution. For all the cases, we provide results for baseline models (using only supervised data), models trained with addition of our TCR over labeled data and finally the models trained using our semi-supervised paradigm. The codes used for our experiments are based on PyTorch. In our experiments, both the supervised and unsupervised data are supplied together in each mini-batch. To maintain their ratio $r$ (see Equation~\ref{eq:supervised} and \ref{eq:unsupervised}), we use separate data loaders for labeled and unlabeled data. 

In the following sections we provide details about the dataset, model and training schemes employed in different image-to-image translation applications. Finally in Sec.~\ref{sec:validation-manifold}, we provide validation for our manifold assumption that is built on the hypothesis that unsupervised data, sampled from the same data distribution as supervised data, can be leveraged to learn a generic image-to-image mapping function.  

\begin{figure*}[t]
    \centering
   {\includegraphics[trim={0cm 5.1cm 6.9cm 0cm}, clip, width=0.95\linewidth]{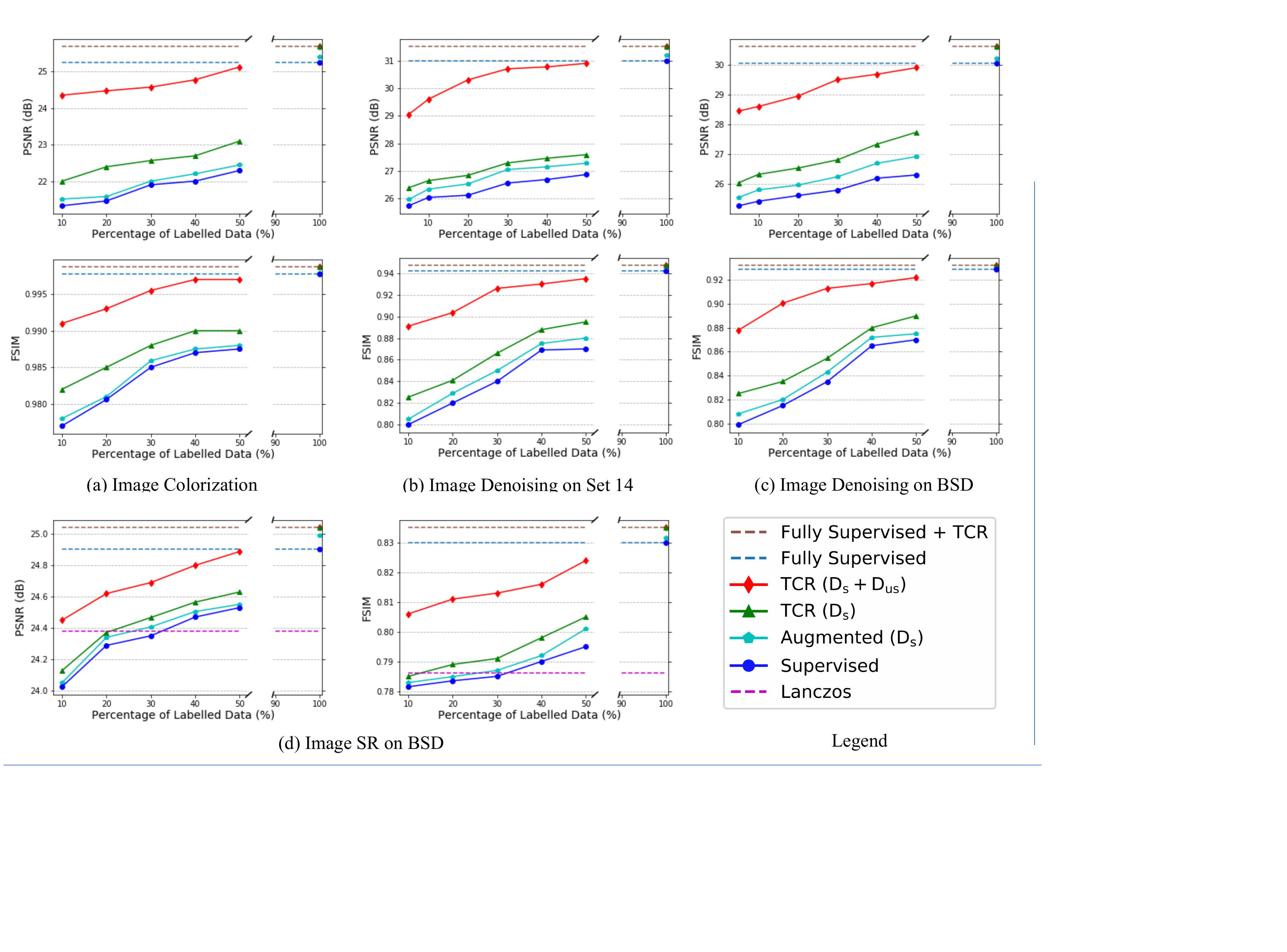} }
    \caption{\small{The plots provide results for baseline models (using only supervised data), models trained with addition of our TCR but using only labeled data i.e. TCR (D$_s$), models trained with image augmentation over the supervised data i.e. Augmented (D$_s$) and finally the models trained using our semi-supervised paradigm i.e. TCR (D$_s$ + D$_{us}$). The addition of unsupervised data while training provides substantial improvement in image reconstruction.}}\label{fig:quantitative-comparison}
    \vspace{-0.3cm}    
\end{figure*}

\subsection{Image Colorization}
\label{sec:image-colorization}
To begin, we compare our semi-supervised method on image colorization. For this, we train the architecture introduced by Iizuka \textit{et al.} \cite{IizukaSIGGRAPH2016} with varying amounts of labeled data (image pairs) from the Places dataset \cite{zhou2017places}. We use the SGD with Nestrov momentum optimizer for all our experiments. The models are trained for 45 epochs with the initial learning rate set to 0.1 with scheduling by a factor of 10 after epoch 30. The momentum parameter was set to 0.9. We achieved the best results when the ratio of supervised to unsupervised data per batch was set at 1:10. Fig.~\ref{fig:qualitative-results}a shows qualitative results for three image samples incrementing the percentage of labeled data used for training the model. In Fig.~\ref{fig:quantitative-comparison}a we compare the performance of our semi-supervised method with the baseline models by means of PSNR and Feature Similarity Index (FSIMc) \cite{zhang2011fsim}. We use FSIMc, rather than SSIM, as it is a more modern metric that was shown to correlate much better with subjective image quality assessment data \cite{Ponomarenko2015} and can operate on both gray-scale and color images.  

\subsection{Image Denoising}
\label{sec:image-denoising}

\begin{figure*}[t]
    \centering
   {\includegraphics[trim={0cm 0cm 0cm 0cm}, clip, width=0.75\linewidth]{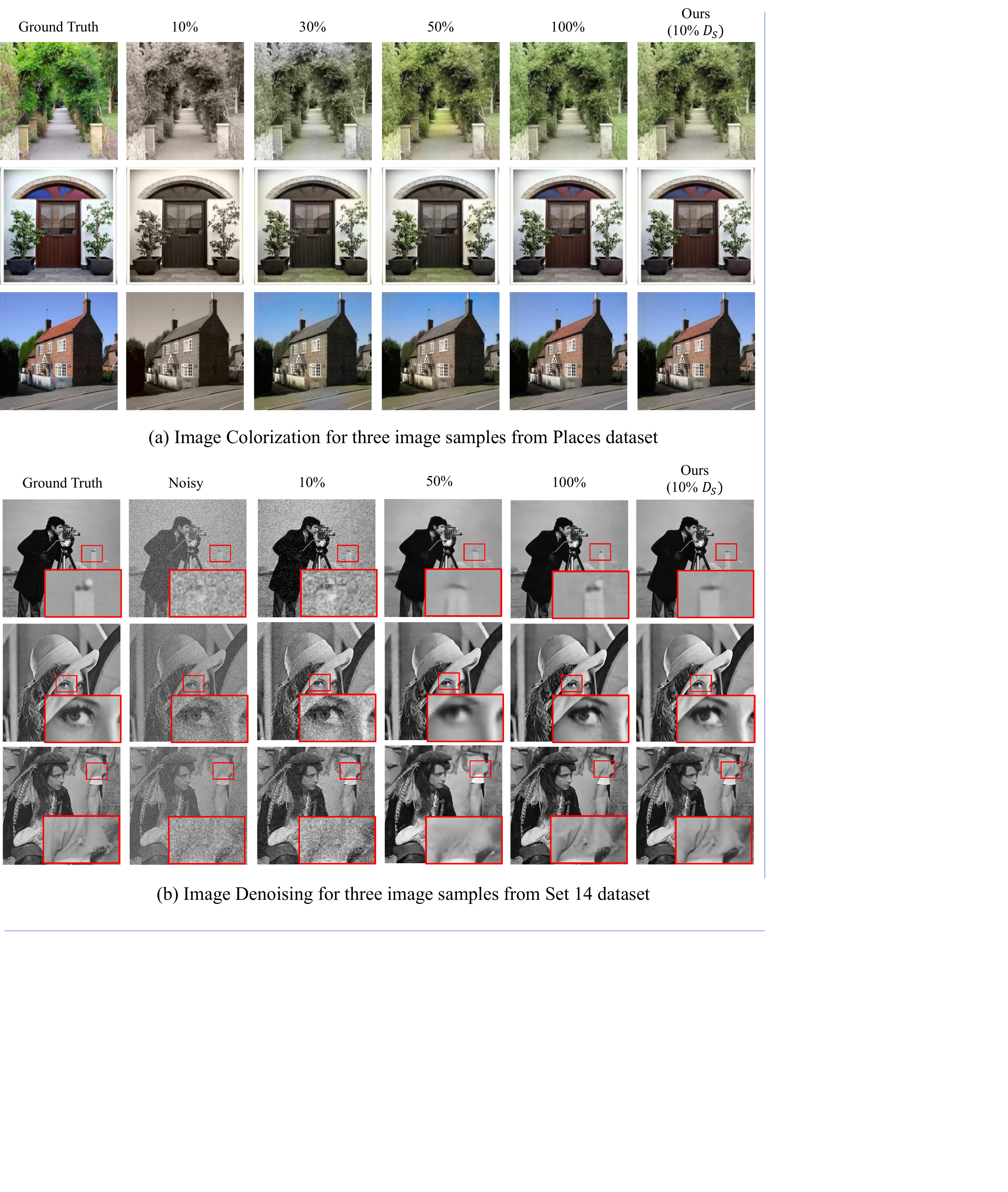} }
    \caption{\small{Qualitative results showing comparison between reconstructed images using our model and supervised baseline models. The column title indicates the percentage of data used for training the model. The last column shows our results where we use only 10\% of the entire dataset as labeled and rest in an unsupervised fashion. Best viewed when zoomed.}}\label{fig:qualitative-results}
    \vspace{-0.5cm}    
\end{figure*}

Next, we test our semi-supervised method on image denoising application. For this, we train the DnCNN architecture introduced by Zhang \textit{et al.} \cite{zhang2017beyond} using the Berkeley Segmentation Dataset (BSD) \cite{MartinFTM01}. The noisy images are generated by adding Gaussian noise with the noise level chosen from the range [0,55]. We use SGD with a weight decay of 0.0001 with Nestrov momentum optimizer for training all our models. The models are trained for 50 epochs with exponential learning rate scheduling from $1e-1$ to $1e-4$. The momentum parameter is set to 0.9. We achieve the best results when the ratio of supervised to unsupervised data per batch was set at 1:10. We choose Set14 data and BSD test data for evaluating of the models efficacy. Fig.~\ref{fig:qualitative-results}b shows qualitative results for three image samples incrementing the percentage of labeled data used for training the model. In Fig.~\ref{fig:quantitative-comparison}b and Fig.~\ref{fig:quantitative-comparison}c we compare the performance of our semi-supervised method with the baseline models on Set 14 and BSD datasets by means of PSNR and FSIM \cite{zhang2011fsim} respectively. 

\subsection{Image Super-Resolution}
\label{sec:image-sr}

Finally, we evaluate the efficacy of our SSL technique on a challenging image-to-image translation method -- Single Image Super-Resolution (SISR). The task of SISR is to estimate a High Resolution (HR) image given a Low Resolution (LR) image. The LR image is first generated by downsampling the original HR image by a factor of $\tau$. For this application we use state-of-the art SISR architecture -- Enhanced Deep Super-Resolution (EDSR) proposed by \cite{lim2017enhanced}, which uses a hierarchy of such residual blocks. We train the models on BSD 500 dataset, choosing 400 images for training and 100 for testing. The upscaling factor $\tau$ is chosen to be 3. Each model is trained for 500 epochs with an initial learning rate of 0.01 with gradual scheduling. We use the PSNR and FSIM \cite{zhang2011fsim} as performance metrics to compare the performance of our SSL scheme with its supervised counterpart (see Fig.~\ref{fig:quantitative-comparison}d). Additional results are included in the appendix.

\subsection{Results and Discussions}
\label{sec:results}

\subsubsection{Comparisons:}

The goal of our TCR is to reconstruct better images using the combination of $D_s$ and $D_{us}$, than what have been obtained using only $D_s$. This is a very common scenario in image-to-image translation as we typically have abundance of input images, but much fewer corresponding (labeled) output images. As illustrated in Fig.~\ref{fig:quantitative-comparison}, we show a substantial performance boost across all application of image translation. We report an absolute PSNR gain of 3.1\,dB and 3.0\,dB for image colorization and denoising using our semi-supervised scheme with only 10\% of labeled data. A similar trend can be observed for FSIM results, suggesting a substantial gain in the perceived quality of translation. We compare the performance of our SSL method for three applications with the supervised baseline, which uses only the labeled data and the same underlying models for training. We further evaluate the performance of our TCR loss term when applied on the labeled data only, as done in \cite{eilertsen2019single}. This is shown as green lines with triangle markers in Fig.~\ref{fig:quantitative-comparison}. We observe that TCR improves the reconstruction quality even when trained only on the labeled data. Finally, we compare all tested approaches to data augmentation, shown as blue lines with pentagon markers in Fig.~\ref{fig:quantitative-comparison}. We used the same set of transformations for data augmentation as for TCR. The results show that image augmentation improves the performance as compared to the baseline models, however, its still inferior to performance achieved using TCR under supervised settings. 

\subsubsection{Movie Applications:}

In this section, we show the potential of our semi-supervised learning method in applications like movie colorization and generation of high resolution of video clips. In movie colorization, we use Blender Foundation's open source short film `Big Buck Bunny' \cite{big-buck-bunny}. We divide the movie into train and test set with each comprising of 510 and 43 seconds respectively. In our SSL setting, we make use of only 1\% of the total training frames in supervised fashion, while rest were fed into the TCR term. We compare our method with its supervised counterpart (using 1\% of total training frames), and achieve an absolute gain of 6.1\,dB in PSNR. Following a similar setting, we further evaluate the efficacy of TCR for denoising and enhancing the resolution of video clips in our experiments.

We believe our semi-supervised learning technique can go a long way in colorization of old movies/video clips requiring an artist to colorize only a few frames. In a similar fashion, we can take advantage of the abundance of unlabeled data in removing noise and enhancing the quality and resolution of movie clips. Additional results and videos are shared in the appendix. 

\vspace{-2.45mm}
\section{Manifold Assumption Validation}
\label{sec:validation-manifold}

In this paper we propose that input image samples (e.g. a noisy image) and their reconstructed counterparts (e.g. denoised image) lie on different manifolds and large portions of unsupervised data can better help in regularizing the remapping function $f_\theta (\cdot)$. Addition of this data while training promotes the reconstructed image to be mapped to the natural image manifold as illustrated in Fig.~\ref{fig:image-mapping}.

To validate this assumption for image denoising, we fine-tune an ImageNet pre-trained Inception-v3 model as a binary classifier using 10,000 pairs of clean (class 0) and noisy (class 1) images, $299 \times 299$ pixels each, drawn from ILSVRC validation set. The noisy samples are generated using two kinds of noise distributions namely, Gaussian and Poisson. For Gaussian noise we randomize the standard deviation $\sigma \in [0,50]$ for each image. For Poisson noise, which is a dominant source of noise in photographs, we randomize the the noise magnitude $\lambda \in [0,50]$. With the learning rate reduced by a factor of 10, we retrain the top-2 blocks while freezing the rest of the network. The global average pooling layer of the model is followed by a batch normalization layer,
drop-out layer and two dense layers (1024 and 1 nodes, respectively). This model now efficiently leverages the subtle difference between clean images and their noisy counterparts and separates the two with a very high accuracy (99.0\%). To validate our manifold assumption we test the denoised images generated using \textit{a)} supervised model using 10\% labeled data and \textit{b)} model trained using TCR with same percentage of labeled data in semi-supervised fashion. The classifier labels only 44.3\% of the denoised images generated using the supervised model as clean, whereas 97.9\% of denoised images generated using our method are labeled as clean. This confirms that vast majority of the images restored using our method are remapped to the clean image manifold.

\begin{figure*}[t]
    \centering
   {\includegraphics[trim={0cm 7.2cm 13cm 0cm}, clip, width=0.95\linewidth]{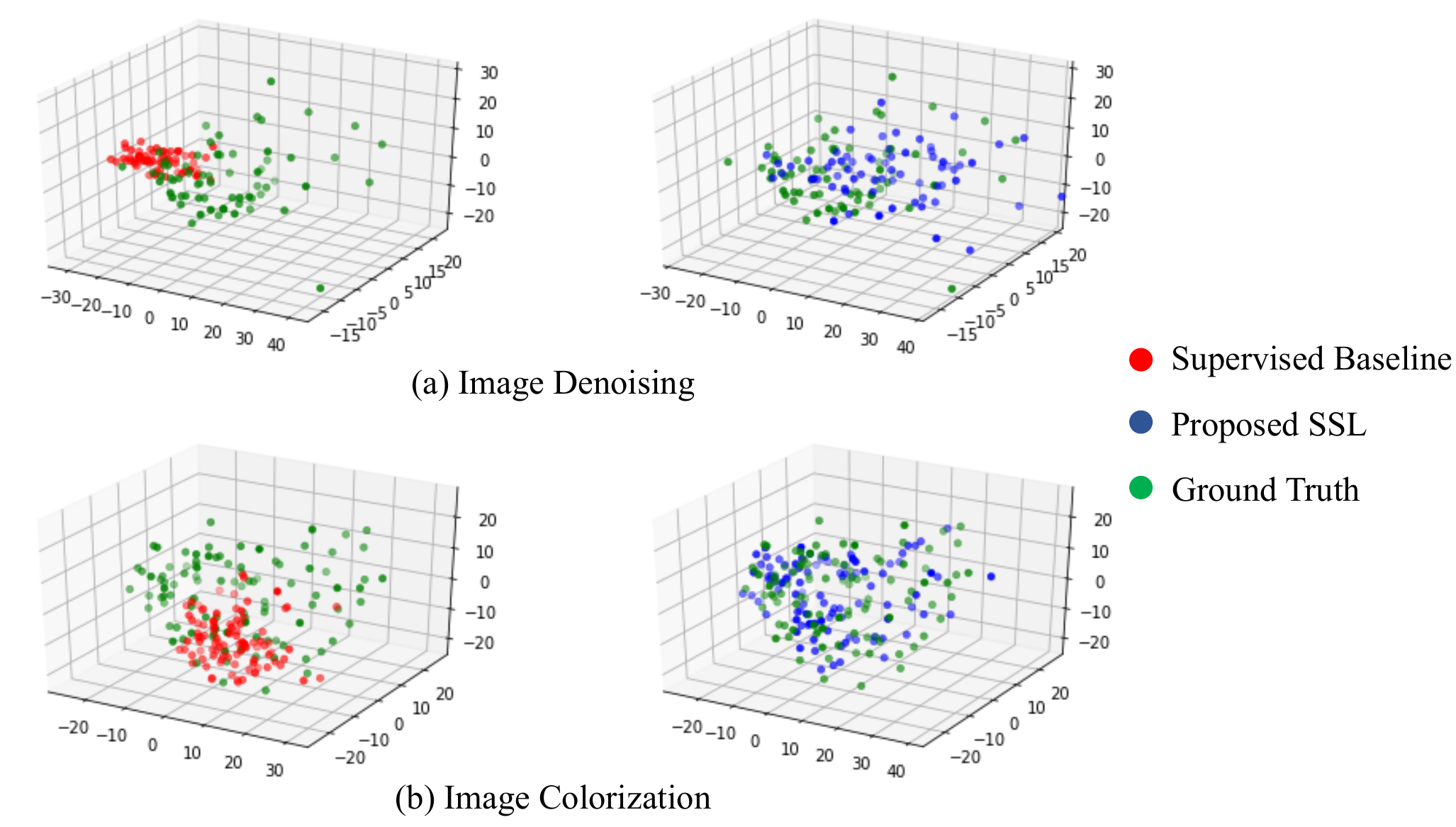} }
    \caption{\small{Manifold Assumption Validation for Image Denoising and Colorization. On the top we show 3D t-SNE plots of the intermediate features extracted from denoised images generated using our semi-supervised model (blue) and the baseline supervised model (red) compared with the ground truth noise-free images (green). The bottom plots show the features extracted from colored images. The plots clearly show that our method provides a more accurate mapping of images back onto the the natural image manifold i.e. the ground truth image manifold than the supervised model.}  }\label{fig:manifold-assumption}
    \vspace{-0.5cm}    
\end{figure*}

Figure.~\ref{fig:manifold-assumption}a, shows a plot of the features extracted from the penultimate layer of the binary classifier to visualize our manifold assumption validation. We employ t-SNE to reduce the  dimensionality of features to 3 for visualization. 

To perform a similar validation of our manifold assumption for image colorization, we use an ResNet-50 model\footnote{https://github.com/CSAILVision/places365} pre-trained on the Places 365 dataset \cite{zhou2017places} to test the efficacy of our colorization model compared to the supervised counterpart. We convert 100 image samples into gray-scale from a particular class of the validation set (houses in our case) and generate colored images using both the aforementioned methods. In Fig.~\ref{fig:manifold-assumption}b, we plot the features extracted after the Global Pooling Layer of the classifier to visualize our manifold assumption validation. We employ t-SNE to reduce the dimensionality of features from 1024 to 3 for visualization. The plots show that TCR helps to project images in each class to the ground truth manifold.

\vspace{-2mm}
\section{Ablation Studies}
\label{sec:ablation-study}

\begin{figure}[t]
    \centering
    \begin{subfigure}[b]{0.488\textwidth}
        \includegraphics[width=\textwidth]{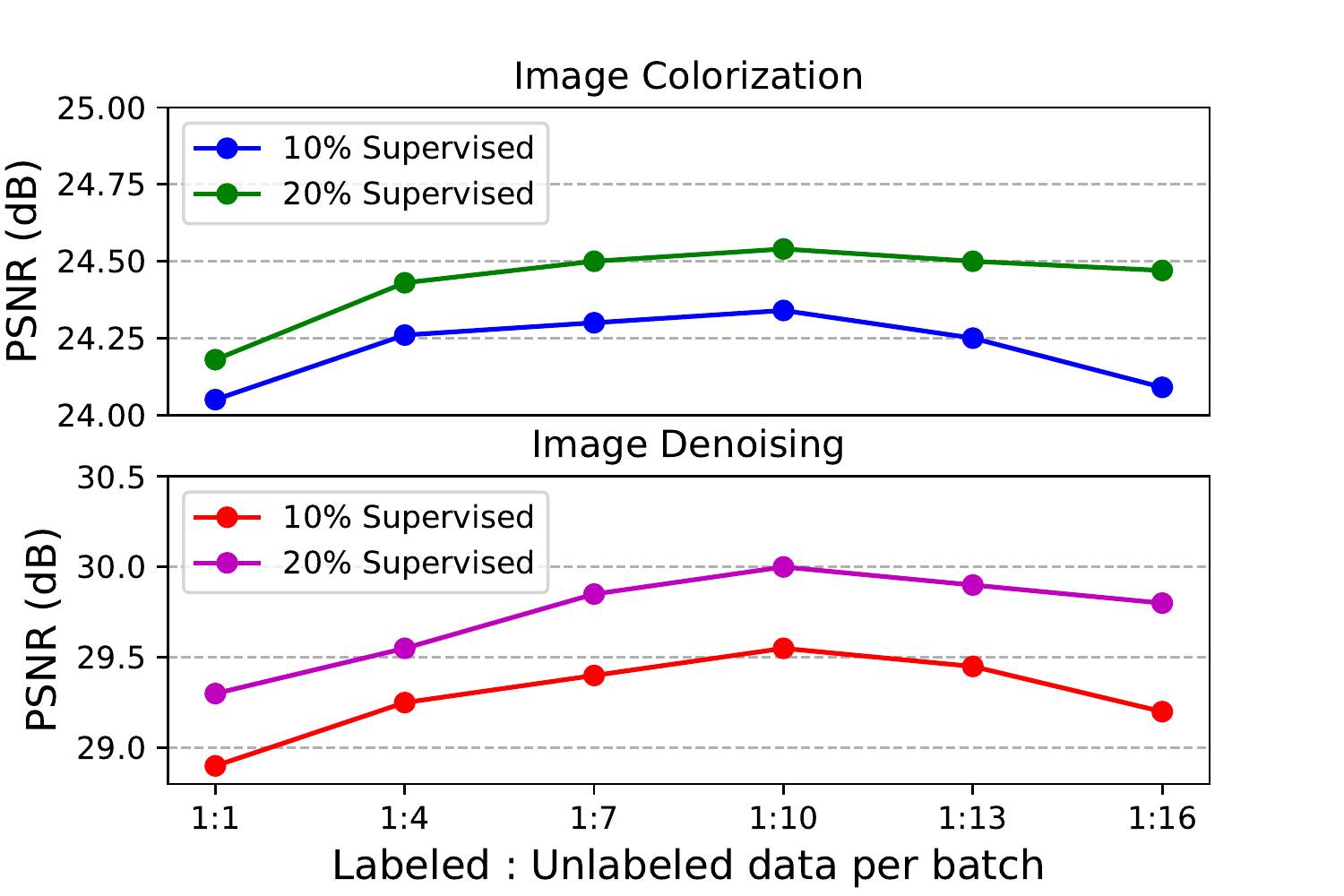}
        \caption{Hyper-parameter $r$.}
        \label{fig:ablation-bs}
    \end{subfigure}
    ~ 
    \begin{subfigure}[b]{0.488\textwidth}
        \includegraphics[width=\textwidth]{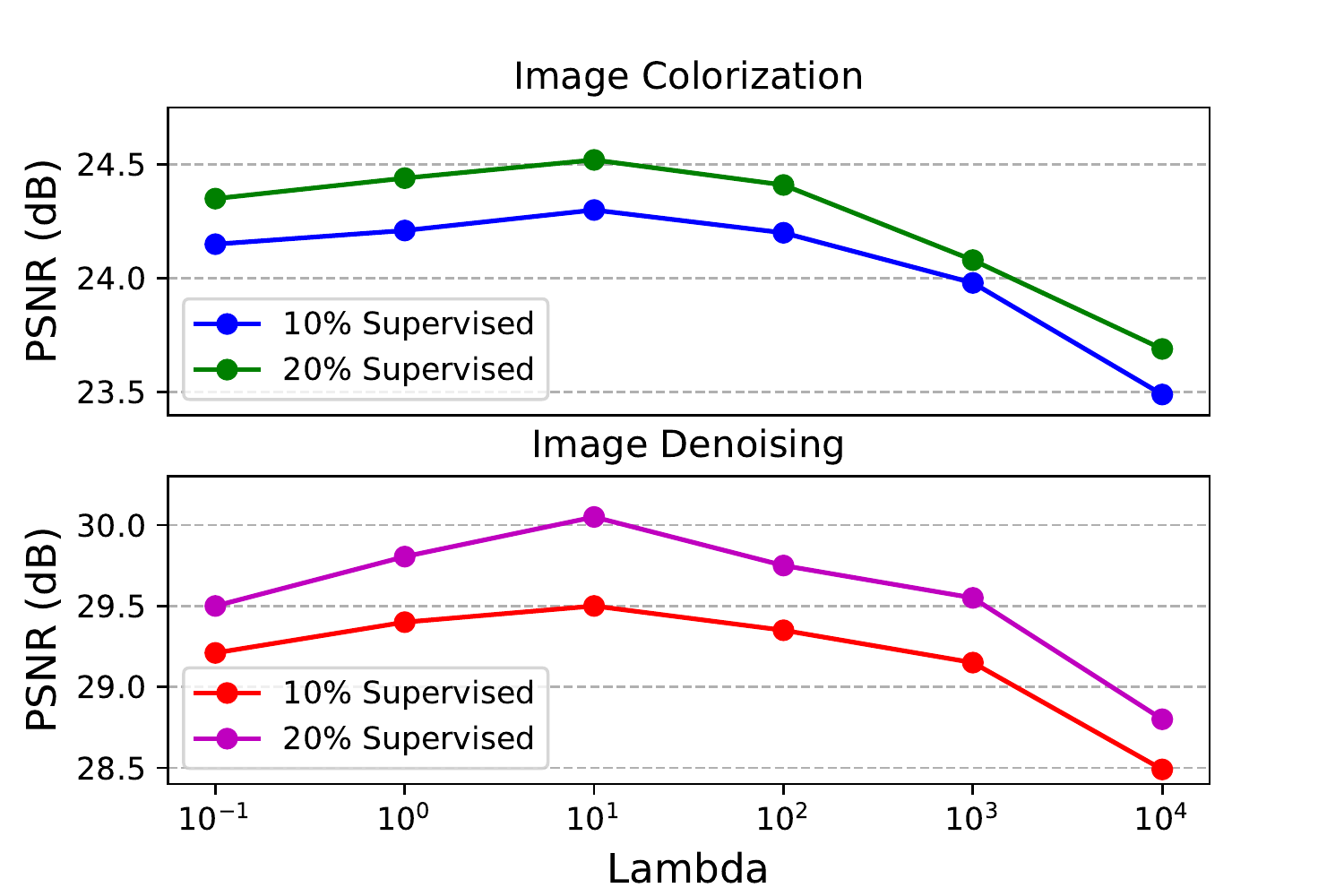}
        \caption{\small{Hyper-parameter $\lambda$.}}
        \label{fig:ablation-lambda}
    \end{subfigure}

    \caption{\small{Ablation Study over two hyper-parameters for image colorization and denoising.}}\label{fig:Ablation-study}
    \vspace{-0.5cm}
\end{figure}

To achieve the best performance, we need to select the values for the hyper-parameters: ratio of unsupervised to supervised data per batch ($r$), and the weight of the regularization term ($\lambda$). We show that the optimal values of both hyper-parameters are consistent across the problems and datasets. 

\subsection{The amount of unsupervised data we need}

First, we study the effect of the ratio of unsupervised to supervised data per batch ($r$) of training in a semi-supervised learning settings. In Fig.~\ref{fig:ablation-bs}, we plot the PSNR values of the reconstructed images versus different ratios for image colorization and denoising. We perform an ablation study using 10\% and 20\% of the total data as labeled. We observe a significant increase in the quality of images by using large amounts of unlabeled data. Our proposed SSL technique has maximum impact across all applications when the the amount of unlabeled data is 10 times that of labeled data in each mini-batch. This helps the model to learn over all possible distributions thereby resulting in a near perfect mapping function using only a small fraction of labeled data. 

\subsection{Lambda}

We conduct a hyper-parameter search over the scalar $\lambda$ used to control the weight given to our unsupervised loss term (see Equation.~\ref{eq:overall}). We searched over values in \{$ 10^{-1}, 10^{0}, 10^{1}, 10^{2}, 10^{3}, 10^{4}$\}. Fig.~\ref{fig:ablation-lambda} plots the PSNR value versus $\lambda$ for image colorization and denoising. Again in the above settings only 10\% and 20\% of the total data is labeled. In our experiments we found the maximum performance boost of our technique for $\lambda = 10$.

\section{Conclusions}
\label{sec:conclusion}
There has been an immense surge in semi-supervised learning techniques, however, none of the methods have addressed image-to-image translation. We introduce Transformation Consistency Regularization, a simple yet effective method that fills in this gap and provides great boosts in performance using only a small percentage of supervised data. The strategy proposes an additional regularization term over unlabeled data and does not require any architectural modifications to the network. We also show the efficacy of our semi-supervised method in colorization, denoising and super-resolution of movie clips by using only a few frames in supervised fashion, while rest been fed to our unsupervised regularization term while training. On the whole, we believe that our method can be used in diverse video applications where knowledge from a few frames can be effectively leveraged to enhance the quality of the rest of the movie.

\section*{Acknowledgements}
This project has received funding from the European Research Council (ERC) under the European Union’s Horizon 2020 research and innovation programme (grant agreement N$^\circ$ 725253–EyeCode).

\newpage

\def\bottomfraction{0.9}

\section*{\Large Appendix}
\setcounter{section}{0}

\section{Performance with Perceptual and $L_1$ Loss}
\label{sec:perceptual-l1}

In our experiments, we follow the exact same training protocol as reported in the original papers for image colorization, denoising ans super-resolution. We found that the addition of Transformation Consistency Regularization (TRC) in a semi-supervised fashion results in substantial performance boost when mean-squared error ($L_2$) is used as a loss function. In this section, we evaluate the efficacy of TRC for image super-resolution application when $L_2$ loss (Eq.~3 in the main paper) is replaced with perceptual loss \cite{gatys2015texture,johnson2016perceptual} or $L_1$ loss \cite{Zhao2017}. 

While pixel level $L_2$ loss is a commonly used SISR protocol, Ledig \textit{et al.} in SRGAN designed a loss function based on the perceptually relevant characteristics of images \cite{ledig2017photo} for image super-resolution. To show TCR is robust to the selection of the loss function, we perform additional experiments relying on using loss function that is closer to the perceptual similarity. For an unlabeled data sample $u_i$ and its geometric transform $T(u_i)$, to compute the perceptual loss, we extract the feature maps from from a pre-trained VGG-19 network $\phi(\cdot)$ \cite{simonyan2014very} and compute the euclidean distance between the two. Mathematically, the perceptual loss $\mathcal{L}_{p}(u)$ is given by:

\begin{equation}
\label{eq:unsupervised-perceptual}
\mathcal{L}_{p}(u)= \frac{1}{rB} \, \sum_{i=1}^{rB} \, \Big( \frac{1}{M} \, \sum_{m=1}^{M} \,{\parallel}\,  T_m \big(\phi(f_\theta (u_i)) \big) - \big(\phi(f_\theta (T_m(u_i)) \big) \,{\parallel}_2^2 \, \Big)
\end{equation}

Here $f_\theta(\cdot)$ is the super-resolution model used in our experiments, $rB$ is the number of unsupervised samples fed to the network per batch and $T_m(\cdot)$ is the $m$-th geometric transformation. To train the model, we equivalently change the supervised loss in Eq.~2 (main paper) to the following:

\begin{equation}
\label{eq:supervised-perceptual}
\mathcal{L}_{s}(x,y)= \frac{1}{B} \, \sum_{i=1}^{B} \, {\parallel} \, \phi \big(f_\theta (x_i) \big) - \phi \big(y_i \big) \,{\parallel}_2^2
\end{equation}

\begin{figure}[t]
    \centering
    \begin{subfigure}[b]{0.488\textwidth}
    \centering
        \includegraphics[width=\textwidth]{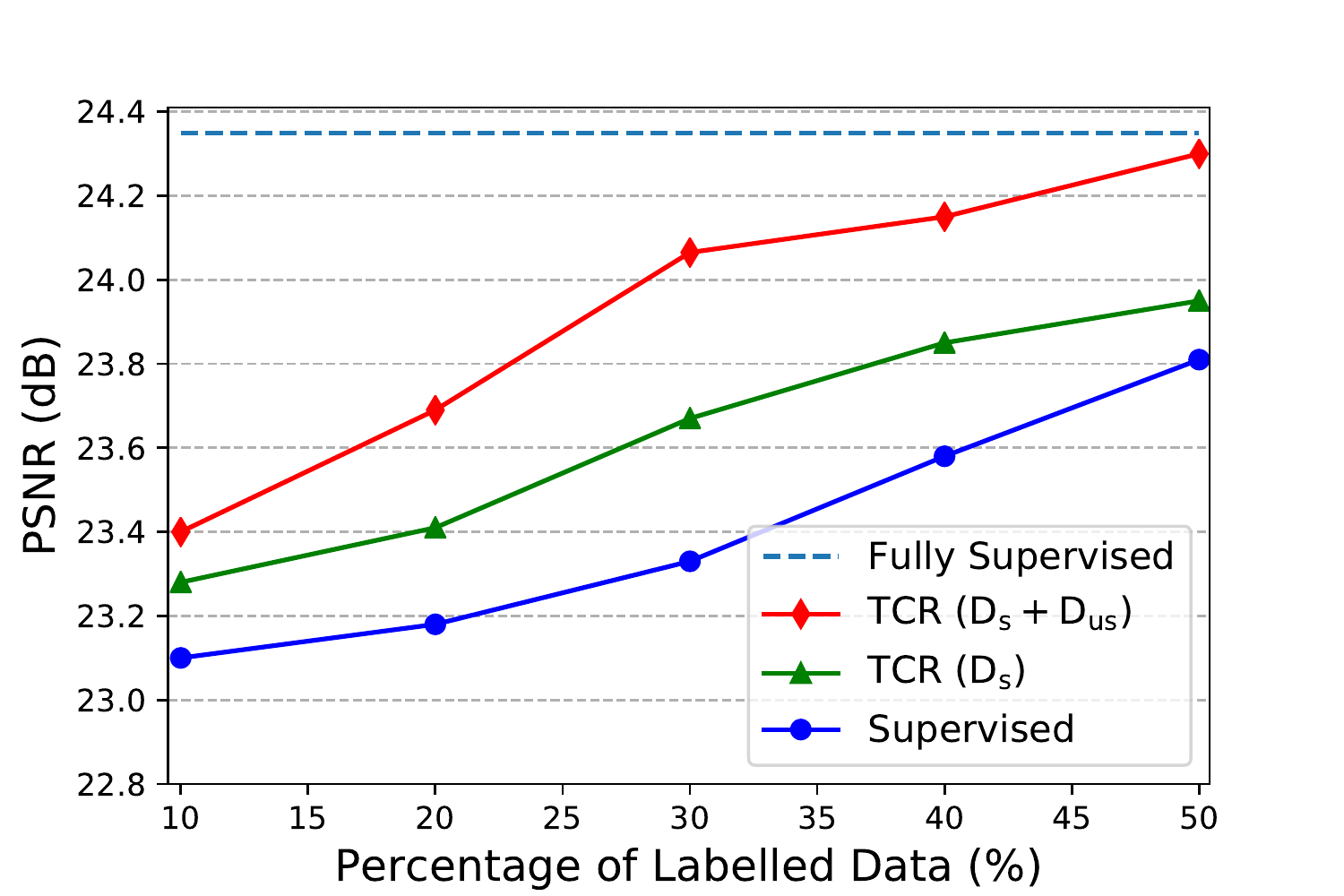}
        \captionsetup{justification=centering}
        \caption{Performance improvement using Perceptual loss for training.}
        \label{fig:perceptual_loss}
    \end{subfigure}
    ~ 
    \begin{subfigure}[b]{0.488\textwidth}
    \centering
        \includegraphics[width=\textwidth]{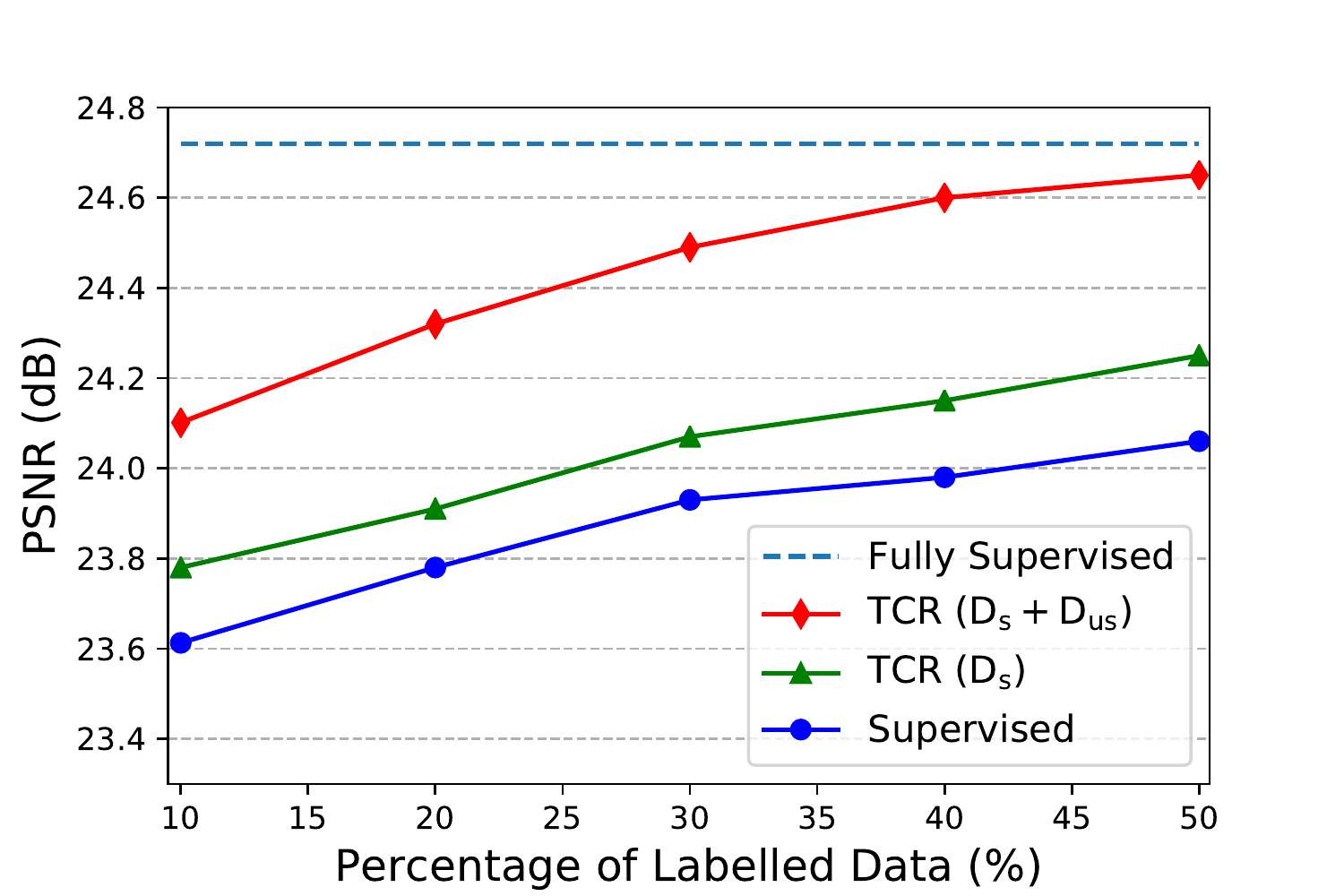}
        \captionsetup{justification=centering}
        \caption{\small{Performance improvement using $L_1$ loss for training.}}
        \label{fig:l1_loss}
    \end{subfigure}

    \caption{\small{The plots shows the efficacy of our method for various types of loss functions for image super-resolution. We show the PSNR values for baseline models (using only supervised data), model trained with addition of our TCR but using only labeled data, and finally models trained using our semi-supervised paradigm. The addition of unsupervised data while training provides substantial improvement in image reconstruction.}}\label{fig:loss_functions}
\end{figure}

The feature representations as extracted from deeper layers of the VGG network, which convey more information about the content of the images \cite{mahendran2016visualizing,gatys2015texture,ledig2017photo}. Fig.~\ref{fig:perceptual_loss} shows a comparison of our semi-supervised scheme using perceptual loss with a baseline model while incrementing the percentage of data used in supervised fashion. Our method consistency results in a significant performance boost compared to its supervised counterpart. Compared to the pixel-level MSE loss, we acheive a lower PSNR value using the perceptual loss, which is consistent with the findings of SR-GAN \cite{ledig2017photo}. 

We further evaluate the performance of TRC using pixel-wise $L_1$ loss for training the image super-resolution model. Fig.~\ref{fig:l1_loss} shows a comparison of our semi-supervised scheme using $L_1$ loss with a baseline model while incrementing the percentage of data used in supervised fashion.

This goes on to show that the proposed semi-supervised method effectively leverages information from unlabeled data performing equally well under various reconstruction loss terms.

\begin{figure*}[t]
    \centering
  {\includegraphics[trim={0cm 0cm 0cm 0cm}, clip, width=0.995\linewidth]{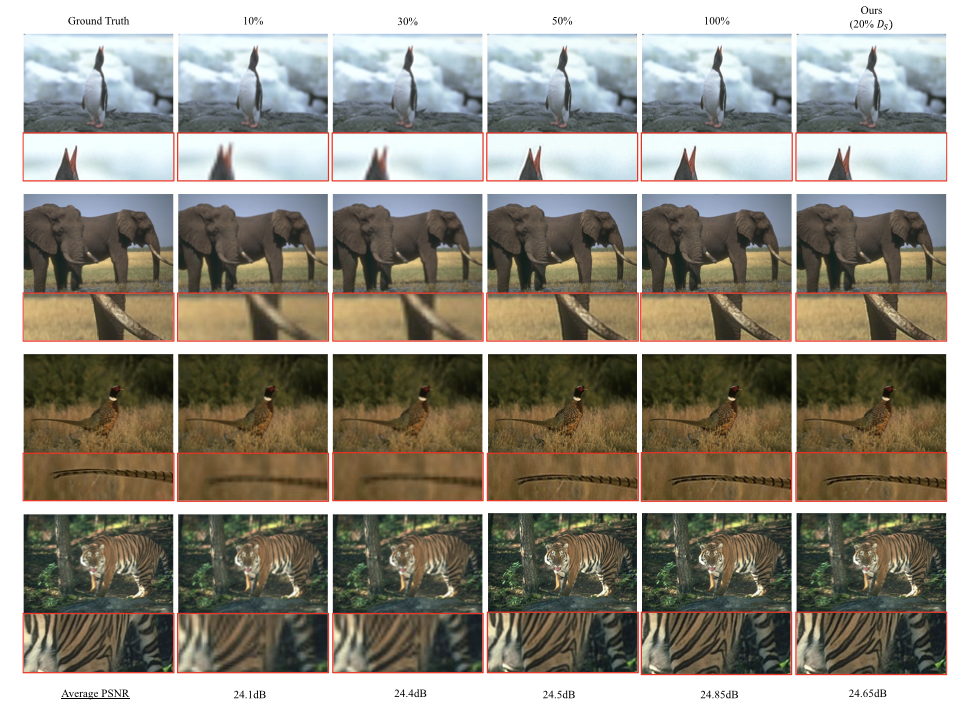}}
    \caption{\small{ Qualitative results showing comparison between reconstructed images using our model and supervised baseline models for single image super-resolution. The column title indicates the percentage of data used for training the model. The last column shows our results where we use only 20\% of the entire dataset as labeled and rest in an unsupervised fashion. Best viewed when zoomed.}}\label{supp-fig:sr_qualitative_figs}
\end{figure*}

\section{Additional Experiments}
\label{sm-sec:additional-experiments}
We further show the potential of our semi-supervised learning method in applications like movie colorization, denoising and generation of high resolution of video clips. 

\begin{figure*}[t]
    \centering
  {\includegraphics[trim={0cm 0cm 0cm 0cm}, clip, width=0.8\linewidth]{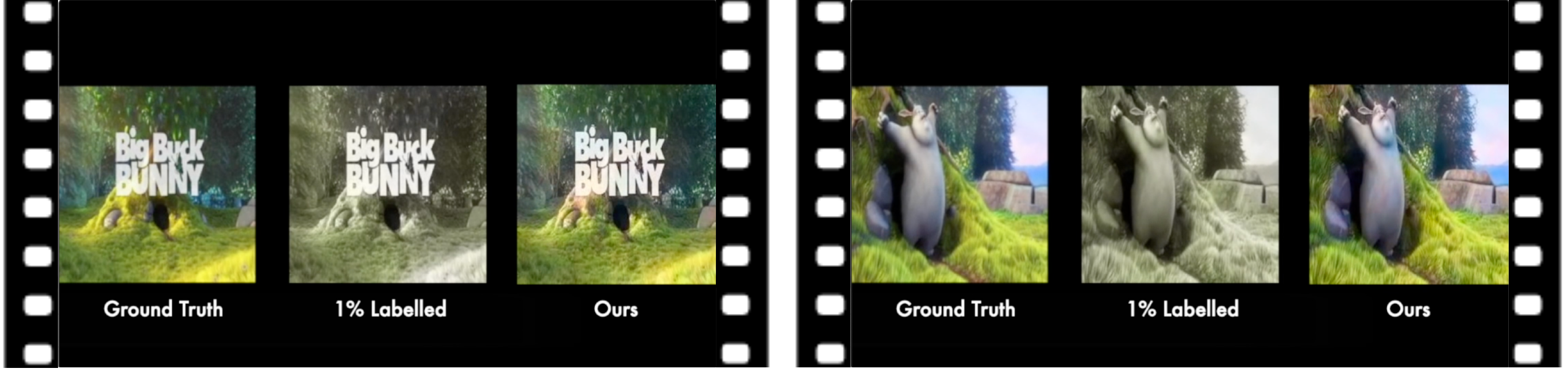}}
    \caption{\small{The figure shows a comparison between our semi-supervised method and the model trained in a completely supervised fashion for colorization. Both the models use only 1\% of the training frames as labeled. Our method results in an average absolute gain of 6.0\,dB compared to the supervised counterpart. Best viewed in color. }}\label{supp-fig:color-movie}
\end{figure*}


\begin{figure*}[h]
    \centering
  {\includegraphics[trim={0cm 0cm 0cm 0cm}, clip, width=0.8\linewidth]{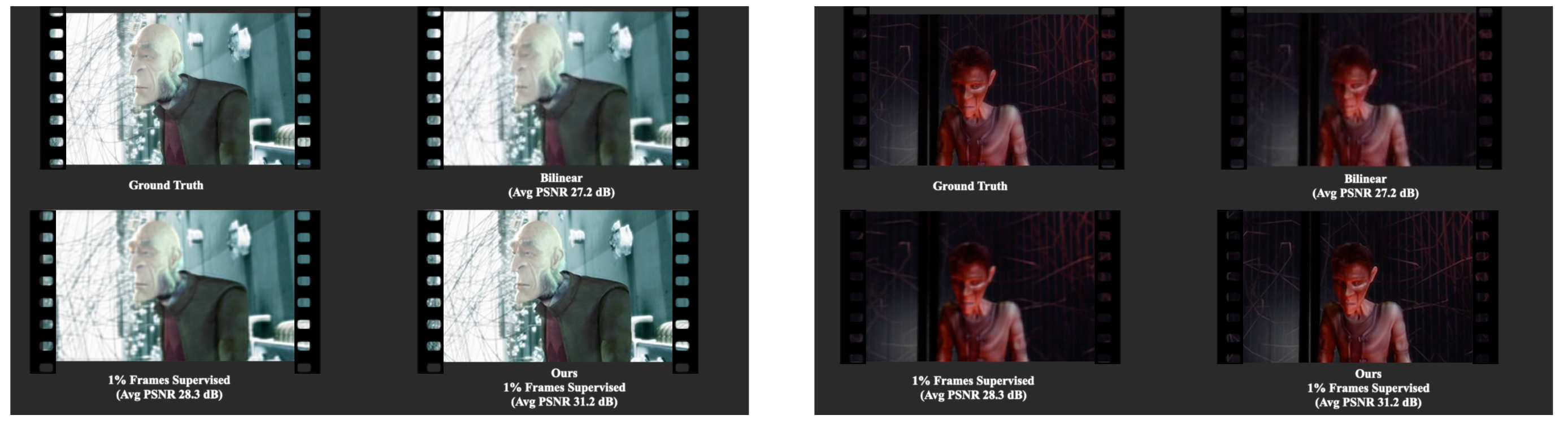}}
    \caption{\small{A snapshot of the video showing comparison between our semi-supervised method and the model trained in a completely supervised fashion for super-resolution (upscale factor $4\times$). Both the models use only 1\% of the training frames as labeled. Our method results in an average absolute gain of 5.7dB compared to the supervised counterpart. Best viewed when zoomed. }}\label{supp-fig:color-sr-4x}
\end{figure*}

\subsection{Movie Colorization}
\label{sm-sec:movie-colorization}

Transformation Consistency Regularization (TRC) can go a long way in colorization of old movies/video clips requiring an artist to colorize only a few frames. To demonstrate the efficacy of our semi-supervised learning (SSL) algorithm for movie colorization we use Blender Foundation's open source short film `Big Buck Bunny' \cite{blender}. We divide the movie into train and test set with each comprising of 510 and 43 seconds respectively. In our SSL settings for image colorization, we made use of only 1\% of the total training frames in supervised fashion, while rest were fed into the TCR term. We achieve an absolute gain of 6.0\,dB in PSNR using our semi-supervised method as compared to the supervised model trained with the same percentage of training data. Fig.~\ref{supp-fig:color-movie} shows a snapshot from the video.

\subsection{Movie Super Resolution}
\label{sm-sec:movie-sr}

Following the same setting, we show how TRC can be used to enhance the resolution of movies by capturing only a few frames at a higher resolution. For this application we use the short film `Elephants Dream' \cite{blender}. The movie frames are divided into train and test sets with each comprising of 600 and 54 seconds respectively. We again use only 1\% of the total training frames in supervised fashion. Our method results in an absolute gain of 5.7\,dB (for up-scaling $4\times$) and 2.9\,dB (for up-scaling $3\times$) compared to the supervised baseline. Fig.~\ref{supp-fig:color-sr-4x}  shows a snapshot from the video for upscale factor $4\times$.

\bibliographystyle{splncs04}
\bibliography{egbib}
\end{document}